\documentclass[10pt,a4paper]{article}
\usepackage{lrec}
\usepackage{times}
\usepackage{url}
\usepackage{latexsym}
\usepackage{cleveref}
\usepackage{adjustbox}
\usepackage{booktabs}
\usepackage{subcaption}

\newcommand{\ignore}[1]{}

\usepackage[disable]{todonotes}

\usepackage{graphicx}
\usepackage{float}

\title{UniMorph 2.0: Universal Morphology}

\name{Christo Kirov$^{1}$, Ryan Cotterell$^{1}$, John Sylak-Glassman$^{1}$, G{\'e}raldine Walther$^2$ \\
\large\bf{Ekaterina Vylomova$^{3}$, Patrick Xia$^{1}$, Manaal Faruqui$^4$, Sabrina J. Mielke$^{1}$, Arya D. McCarthy$^{1}$} \\
\large\bf{Sandra K\"ubler$^5$, David Yarowsky$^{1}$, Jason Eisner$^{1}$, Mans Hulden$^{6}$}}

\address{$^{1}$Johns Hopkins University, $^2$University of Zurich, $^3$University of Melbourne\\
 $^4$Google, $^5$Indiana University, $^6$University of Colorado \\
         Baltimore, Zurich, Melbourne, New York, Bloomington, Boulder \\
         \{ckirov1, ryan.cotterell, jcsg, paxia, sjmielke, arya, yarowsky, eisner\}@jhu.edu, geraldine.walther@uzh.ch\\
         evylomova@gmail.com, mfaruqui@google.com, skuebler@indiana.edu, mans.hulden@colorado.edu\\}
         
\abstract{
The Universal Morphology (UniMorph) project is a collaborative effort to improve how NLP handles complex morphology across the world's languages. The project releases annotated morphological data using a universal tagset, the UniMorph schema. Each inflected form is associated with a lemma, which typically carries its underlying lexical meaning, and a bundle of morphological features from our schema. Additional supporting data and tools are also released on a per-language basis when available. UniMorph is based at the Center for Language and Speech Processing (CLSP) at Johns Hopkins University in Baltimore, Maryland. 
This paper details advances made to the collection, annotation, and dissemination of project resources since the initial UniMorph release described at LREC 2016. \\ \newline  }

\begin{document}

\maketitleabstract


\section{Introduction}

Complex morphology is ubiquitous among the languages of the world. For example, roughly 80\% of languages use morphology to mark verbal tense and 65\% mark nominal case \cite{haspelmath2005world}. While overlooked in the past, explicit modeling of morphology has been shown to improve performance on a number of downstream HLT tasks, including including machine translation (MT) \cite{dyer2008generalizing}, speech recognition \cite{creutz2007analysis}, parsing \cite{TACL631}, keyword spotting \cite{narasimhan2014morphological}, and word embedding \cite{CotterellSE16}. This has led to a surge of new interest and work in this area \cite{durrett2013supervised,ahlberg2014semi,nicolai2015inflection,faruqui2015morphological}.

The Universal Morphology (UniMorph) project, centered at the Center for Language and Speech Processing (CLSP) at Johns Hopkins University 
is a collaborative effort to improve how NLP systems handle complex morphology across the world's languages. The project releases annotated morphological data using a universal tagset, the UniMorph schema. Each inflected form is associated with a lemma, which typically carries its underlying lexical meaning, and a bundle of morphological features from our schema.  Additional supporting data and tools are also released on a per-language basis when available.

\newcite{kirovsylak-glassman2016} introduced version 1.0 of the UniMorph morphological database, created by extracting and normalizing the inflectional paradigms included in Wiktionary (\url{www.wiktionary.org}), a large, broadly multi-lingual crowd-sourced collection of lexical data. This paper describes UniMorph 2.0. It details improvements in Wiktionary extraction and annotation, as well as normalization of non-Wiktionary resources, leading to a much higher quality morphological database. The new dataset spans 52 languages representing a range of language families. As in UniMorph 1.0, we provide paradigms from highly-inflected open-class word categories --- nouns, verbs, and adjectives. Many of the included languages are extremely low-resource, e.g., Quechua, Navajo, and Haida. This data was used as the basis for the CoNLL 2017 Shared Task on Morphological Learning (\url{http://sigmorphon.org/conll2017}) \cite{cotterell-conll-sigmorphon2017}.

\section{Wiktionary Extraction}

\begin{figure*}[!ht]
\centering
\begin{adjustbox}{width=1\textwidth}
\includegraphics[scale=0.65]{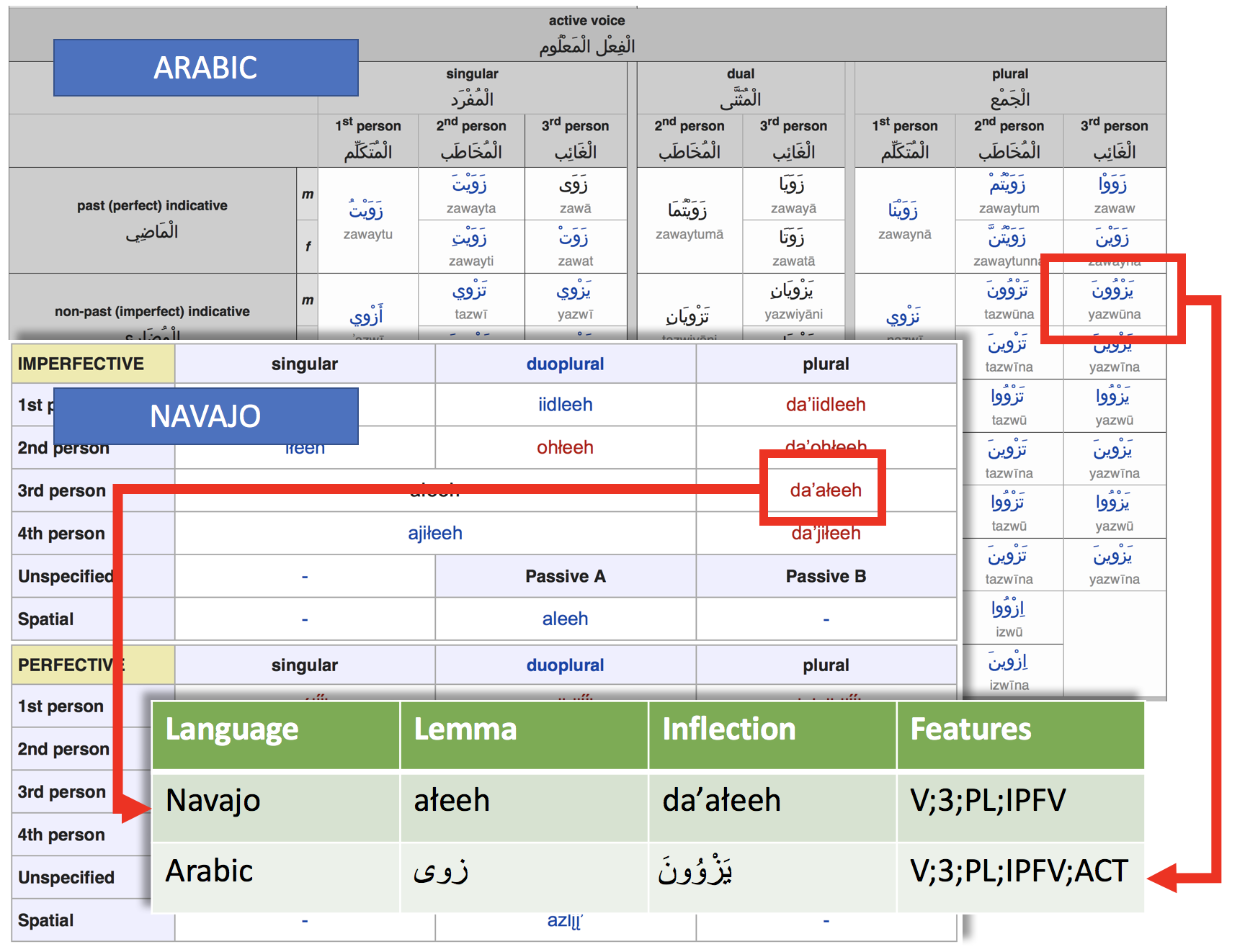}
\end{adjustbox}
\caption{\label{fig:challenge}
Paradigm extraction and normalization.}
\end{figure*}

In \newcite{kirovsylak-glassman2016}, we introduced version 1.0 of the UniMorph morphological database, based on a very large-scale parsing and normalization of Wiktionary. Wiktionary is a broadly multilingual resource with many crowd-sourced morphological paradigms in the form of custom HTML tables. Figure~\ref{fig:challenge} illustrates the challenge associated with extracting this data. Wiktionary is designed for human, rather than machine readability, and authors have extensive freedom in formatting data. This leads to wildly differing table layouts across languages which need to be converted to a consistent tabular format.

The extraction process developed for UniMorph 1.0 relied heavily on statistical, visual, and positional heuristics \cite{sylak-glassmankirov2015a} to:
\begin{enumerate}
\item Determine which entries in an HTML table are inflected forms and which are grammatical descriptors.
\item Link each inflected form with its appropriate descriptors.
\item Convert each set of linked descriptors into a universal feature annotation schema, described in detail in Sylak-Glassman (2016).\footnote{\url{unimorph.github.io/doc/unimorph-schema.pdf}}
\end{enumerate}
This led to a large dataset of 952,530 unique noun, verb, and adjective lemmas across 350 languages. Unfortunately, the UniMorph 1.0 dataset was very error-prone due to the inability of our heuristics to fully cover the degree of inconsistency found in Wiktionary. For many lemmas, inflected forms were systematically linked to incorrect feature vectors. To correct these errors, we noted that for each part-of-speech within a language in Wiktionary, authors use only a handful of distinct table layouts. Thus, it was sufficient for a human to verify and correct a single lemma parse from a particular layout, and apply those corrections to all similar lemma parses. A custom verification and correction process was created and applied to 8 languages (Arabic, Finnish, Georgian, German, Navajo, Russian, Spanish, and Turkish) in preparation for the SIGMORPHON 2016 Shared Task on Morphological Reinflection \cite{cotterellsigmorphon2016}.

\begin{figure}
    \centering
    \begin{subfigure}[b]{0.7\columnwidth}
        \includegraphics[width=\textwidth]{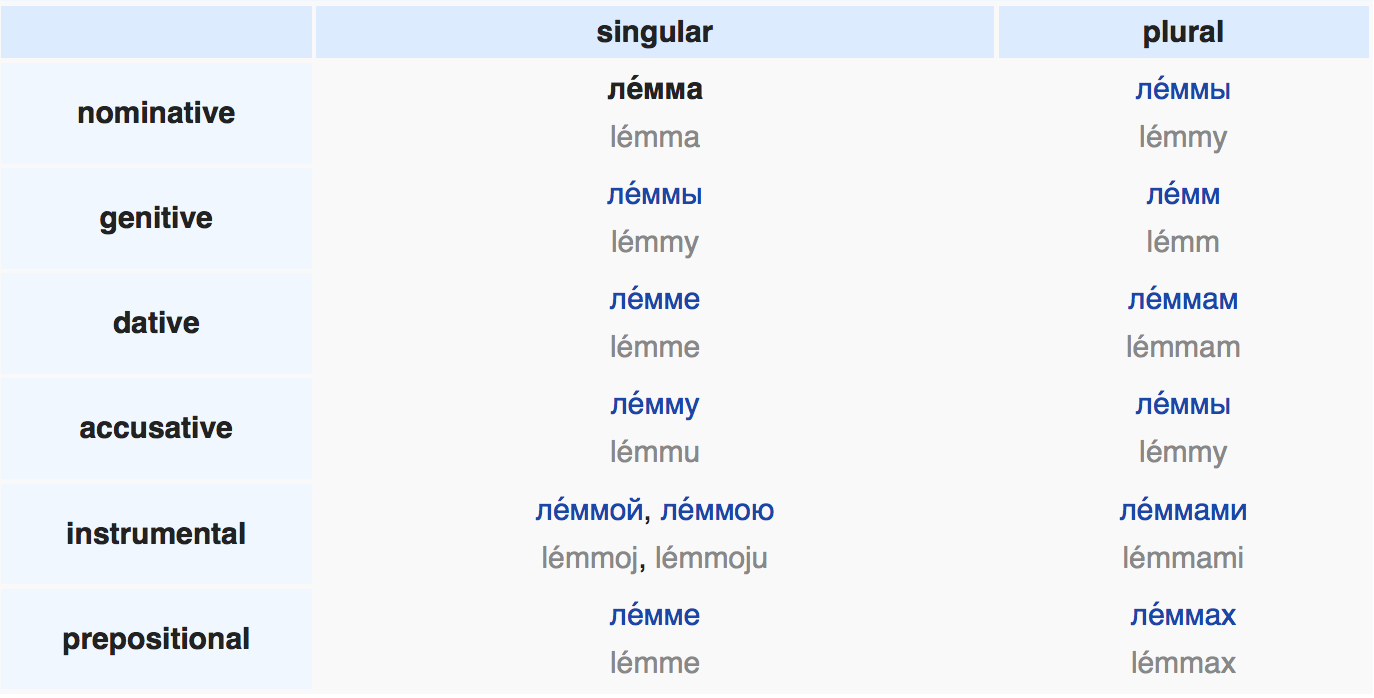}
        \caption{Raw Wiktionary}
        \label{fig:gull}
    \end{subfigure}
    ~ 
    \begin{subfigure}[b]{0.7\columnwidth}
        \includegraphics[width=\textwidth]{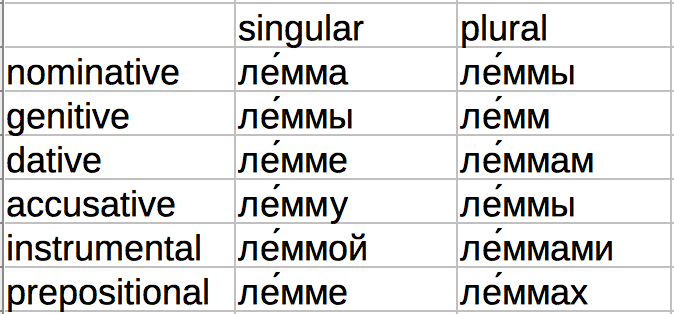}
        \caption{Unannotated Table}
        \label{fig:tiger}
    \end{subfigure}
    ~ 
    \begin{subfigure}[b]{0.7\columnwidth}
        \includegraphics[width=\textwidth]{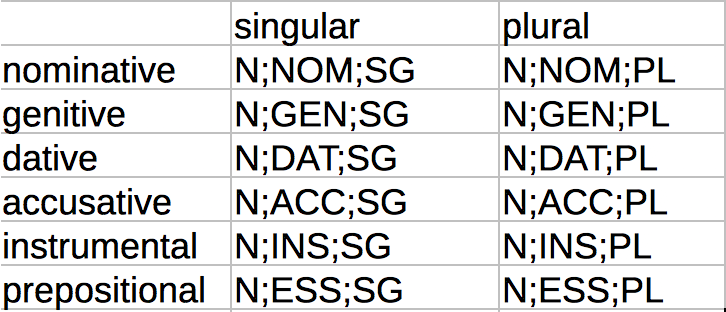}
        \caption{Annotated Table}
        \label{fig:mouse}
    \end{subfigure}
    \caption{Annotation process}\label{fig:annotation}
\end{figure}

For UniMorph 2.0, we noticed that the effort required to verify and correct an automatic parse of a Wiktionary table layout was greater than or equal to the effort required for a human to \emph{directly} annotate a table with UniMorph features instead. Figure~\ref{fig:annotation} illustrates this simplified process. Each language's HTML tables were parsed using Python's \texttt{pandas} library (\url{pandas.pydata.org}) and grouped according to their tabular
structure and number of cells.  Each group represents a different type of paradigm (e.g., regular verb). 

For each group, a sample table was selected, and an annotator replaced each inflected form in the table with the appropriate UniMorph features. All annotation was compliant with the UniMorph Schema, which was designed to represent the full range of semantic distinctions that can be captured by inflectional morphology in any language \cite{sylak-glassmankirov2015}. The schema is similar in form and spirit to other tagset universalization efforts, such as the Universal Dependencies Project \cite{choimarneffe2015} and Interset \cite{zeman2008}, but is designed specifically for typological completeness for inflectional morphology, including a focus on the morphology of especially low-resource languages. It includes over 200 base features distributed among 23 dimensions of meaning (i.e., morphological categories), including both common dimensions like tense and aspect as well as rarer dimensions like evidentiality and switch-reference. Despite the high coverage of the UniMorph tagset, for UniMorph 2.0, annotators were allowed to employ additional `language specific' LGSPEC(1, 2, 3, etc.)  features to mark any missing distinctions, or purely optional form variants that are not associated with a semantic difference. The Spanish imperfect subjunctive, for example, has two interchangeable forms (\emph{-ra} and \emph{-se}):

\vspace{+.5em}
\begin{tabular}{l l l}
\toprule
\emph{lemma} & form & features\\
\midrule
\emph{gravitar}& gravitaras & V;SBJV;PST;2;SG;LGSPEC1\\
\emph{gravitar}& gravitases & V;SBJV;PST;2;SG;LGSPEC2\\
\bottomrule
\end{tabular}
\vspace{+.5em}

As each example table is identical in structure to all members in the same layout group, annotating just one example allows mapping every inflected form in every table in the group to its corresponding morphological features. This minimizes the human annotation effort required per language, to the point that only 3 annotators were able to produce a complete initial dataset for 47 Wiktionary languages in a matter of days (data for these 47 languages, listed in Table~\ref{tab:dq}, supplants the corresponding language data in the UniMorph 1.0 dataset). 

Some of the extracted paradigms from Wiktionary were subject to additional post-processing. In particular, some Wiktionary tables contain multiple forms in the same cell. In the case of multiple forms, we separated them into their own entries. Looking at another Spanish example, we separate \emph{tu} and \emph{vos} forms corresponding to dialect differences in the choice of second person pronoun.

\vspace{+.5em}
\begin{tabular}{l l l}
\toprule
\emph{gravitar}	& gravitas(t\'u) 	&V;IND;PRS;2;SG\\
	& gravit\'as(vos) & \\
\midrule
\emph{gravitar}& gravitas & V;SBJV;PST;2;SG;LGSPEC3\\
\emph{gravitar}& gravit\'as & V;SBJV;PST;2;SG;LGSPEC4\\
\bottomrule
\end{tabular}
\vspace{+.5em}

Finally, the content of all initial annotations was also verified as linguistically sensible by a second, larger set of adjudicators who were either native speakers of the language they reviewed or had significant expertise through research. The final dataset sizes are given by language in \cref{tab:dq}.

\section{Non-Wiktionary Data Sources}

In additional to our large database of 
annotated inflected forms derived from Wiktionary, UniMorph 2.0 includes morphological data for several additional languages from non-Wiktionary sources. Data for
Khaling, Kurmanji Kurdish, and Sorani Kurdish was derived from 
the Alexina project
\cite{waltherjacques2013,walthersagot2010b,walthersagot2010}.\footnote{\url{https://gforge.inria.fr/projects/alexina/}}
Novel data for Haida, a severely endangered North American language isolate, was prepared by Jordan Lachler (University of Alberta). 
Basque language data was extracted from a manually designed
finite-state morphological analyzer \cite{alegria2009}. Data for all these additional languages was reformatted to match the Wiktionary-derived data using custom Python scripts. Any dataset-specific annotation was manually mapped to the UniMorph schema standard.

\section{Supplementary Structured Data}


As discussed in \newcite{kirovsylak-glassman2016}, we also mine additional structured data from Wiktionary. A number of Wiktionary pages contain lists of derived words under the HTML heading `Related/Derived Terms' --- `sunflower' for example, appears on the list for the base lemma `flower.' Furthermore, Wiktionary also contains tables of lemma translations. The English lemma `flower' contains the translation entry `Danish: blomstre.' As part of UniMorph 1.0, we collected an average of 3.42 derived terms per lemma across 76,038 lemmas, and an average of 3.54 translations per annotated lemma.

For UniMorph 2.0, we are releasing two additional resource types. First, only a subset of Wiktionary languages and lemmas contain embedded morphological tables. There are many more bare lemmas with no form of morphological annotation. We also scrape these lemmas, and provide a list of them along with their associated part of speech.

Second, for a number of languages in UniMorph, we provide multi-word English glosses for complex inflected wordforms. For example, the Spanish word \textit{compr\'abamos} is mapped to the gloss `(we) were buying.' These glosses are generated for languages where adequately-sized lemma-to-lemma translation dictionaries are available, via the following general process:
\begin{enumerate}
\item Perform a generally language-independent conversion of UniMorph feature vectors to an English gloss template, e.g., V;1;PL;PST;IPFV $\rightarrow$ `(we) were VBG.' Here, VBG is a Penn Treebank tag which indicates that the template can be filled with the \textit{-ing} form of an English verb.
\item Given an inflected lemma in the language with a particular feature vector and lemma translation, find the corresponding gloss template, e.g., \textit{compr\'abamos}, \textit{comprar}, V;1;PL;IPFV $\rightarrow$ `buy: (we) were VBG'
\item Replace the English lemma placeholder in the template with the appropriately generated form of the English lemma, `buy, (we) were VBG' $\rightarrow$ `(we) were buying'
\end{enumerate}

Generating complicated tenses of multi-word lemmata (e.g. ``They will not have looked it up'') and robustly generating appropriate English inflections for diverse and noisy translation dictionaries, are both a challenge and strength of this work.

Table~\ref{tab:glosses} shows the a summary of the current resource sizes of selected languages, along with the number of distinct inflections covered, and the number of expanded phrasal glosses generated given multiple translations per lemma.

\begin{table}
\centering
\begin{tabular}{l|r|r}
\toprule
\textbf{Language} & \textbf{Inflections} & \textbf{Glosses} \\
\midrule
Amharic & 566553 & 1736981 \\
Farsi & 206711 & 582449 \\
Hausa & 55860 & 124492 \\
Hungarian & 2814006& 9754197 \\
Oromo & 26690 & 246856 \\
Russian & 560067 & 2219960 \\
Somali & 451217 & 1144096 \\
Spanish & 153121 & 368636 \\
Ukrainian & 20288 & 41590 \\
Yoruba & 127833 & 356502 \\
\midrule
{\bf Total} & {\bf 4982569} & {\bf 16575759} \\
\bottomrule
\end{tabular}
\caption{English glosses by language.}
\label{tab:glosses}
\end{table}

\begin{table}
\centering
\begin{adjustbox}{width=.9\columnwidth}
\begin{tabular}{l | l | r@{\ /\ } l}
\toprule
\textbf{Language} & \textbf{Family} & \textbf{Lemmata}&\textbf{Forms} \\
\midrule
Albanian & Indo-European &589&33483  \\
Arabic & Semitic&4134&140003   \\
Armenian &Indo-European&7033&338461   \\
Basque & Isolate &26&11889 \\
Bengali & Indo-Aryan&136&4443   \\
Bulgarian & Slavic&2468&55730    \\
Catalan & Romance&1547&81576 \\
Czech & Slavic&5125&134527  \\
Danish &Germanic&3193&25503    \\
Dutch &Germanic&4993&55467  \\
English &Germanic&22765&115523    \\
Estonian &Uralic&886&38215 \\
Faroese & Germanic&3077&45474 \\
Finnish & Uralic&57642&2490377 \\
French &Romance&7535&367732   \\
Georgian &Kartvelian&3782&74412   \\
German &Germanic&15060&179339    \\
Haida &Isolate&41&7040    \\
Hebrew &Semitic&510&13818 \\
Hindi &Indo-Aryan&258&54438  \\
Hungarian &Uralic&13989&490394 \\
Icelandic &Germanic&4775&76915  \\
Irish &Celtic&7464&107298 \\
Italian &Romance&10009&509574 \\
Khaling &Sino-Tibetan&591&156097  \\
Kurmanji Kurdish &Iranian&15083&216370  \\
Latin &Romance&17214&509182 \\
Latvian &Baltic&7548&136998   \\
Lithuanian &Baltic&1458&34130  \\
Lower Sorbian &Germanic &994&20121 \\
Macedonian &Slavic&10313&168057   \\
Navajo &Athabaskan&674&12354 \\
Northern Sami &Uralic&2103&62677   \\
Norwegian Bokm{\aa}l &Germanic&5527&19238   \\
Norwegian Nynorsk & Germanic&4689&15319    \\
Persian &Iranian&273&37128   \\
Polish &Slavic&10185&201024   \\
Portuguese & Romance&4001&303996    \\
Quechua &Quechuan&1006&180004    \\
Romanian &Romance&4405&80266  \\
Russian &Slavic&28068&473481   \\
Scottish Gaelic&Celtic&73&781   \\
Serbo-Croatian & Slavic&24419&840799   \\
Slovak &Slavic&1046&14796   \\
Slovene &Slavic&2535&60110    \\
Sorani Kurdish &Iranian&274&22990  \\
Spanish &Romance&5460&382955   \\
Swedish &Germanic&10553&78411    \\
Turkish &Turkic&3579&275460    \\
Ukrainian &Slavic&1493&20904   \\
Urdu &Indo-Aryan&182&12572    \\
Welsh &Celtic&183&10641    \\
\bottomrule
\end{tabular}
\end{adjustbox}
\caption{Total number of lemmata and forms available for each language in the morphological database.}
\label{tab:dq}
\end{table}

\section{Community Features}

Following the model of Universal Dependencies (UD),\footnote{\url{universaldependencies.org}}, UniMorph is intended to be a highly collaborative project. To that end, all data and tools associated with the project are released on a rolling basis with a permissive open source license. The main portal for the UniMorph project, which provides a high-level overview of project goals and activities, is \texttt{www.unimorph.org}. The hub for downloadable data and resources is \texttt{unimorph.github.io}. A full specification of the UniMorph annotation schema is available. For each language, the site indicates how many forms and paradigms have been extracted, the source of the data, and available parts of speech. The site is also designed to encourage community involvement. Each language is associated with a public issue tracker that allows users to discuss errors and issues in the available data and annotations. Interested users can also become part of the UniMorph mailing list.

Moving forward, we also intend to develop connections with other morphological resources. The Universal Dependencies project, for example, provides a token-level corpus complementary to the UniMorph type-level data. A preliminary survey of UD annotations shows that approximately 68\% of UD features map directly to UniMorph schema equivalents. This set covers 97.04\% of complete UD tags. Some UD features lie outside the current scope of UniMorph, which marks primarily morphosyntactic and morphosemantic distinctions. These include, for example, markers for abbreviated forms and foreign borrowings.

\section{Conclusion}

As part of the UniMorph project, we are releasing the largest available database of high-quality morphological paradigms across a typologically-diverse set of languages. To create this dataset, we developed a type-based annotation procedure that enables extracting a large amount of data from Wiktionary with minimal effort from human annotators. The procedure successfully handles idiosyncratic variation in formatting across the languages in Wiktionary. UniMorph also prescribes a universal tagging schema and data formats that allow  data to be incorporated from non-Wiktionary data sources. The project welcomes community involvement, and all data and tools are released under a permissive open-source license at \url{unimorph.github.io}. UniMorph 2.0 data has already been used as the basis for the successful CoNLL 2017 Shared Task on Morphological Learning, the first shared task on morphology in the CoNLL community \cite{cotterell-conll-sigmorphon2017}. 



\Urlmuskip=0mu plus 1mu\relax
\nocite{chelliahreuse2011,bickelnichols2005}
\bibliographystyle{lrec}

\section{Bibliographical References}

\bibliography{conll_proposal}

\end{document}